\documentclass[review,a4paper]{cas-sc}
\usepackage{amsmath,amssymb,amsfonts}
\usepackage{graphicx,subfigure}
\usepackage{caption,subcaption}
\usepackage{bm}
\usepackage{xurl}
\usepackage{balance}
\usepackage{autobreak}
\usepackage[authoryear]{natbib}
\usepackage[english]{babel} % for hyphenation

\usepackage[printonlyused]{acronym}
    \acrodef{BNIOA}{Bio-/Nature-inspired optimization algorithm}
    \acrodef{STOA}{Sooty Tern Optimization Algorithm}
    \acrodef{TSA}{Tunicate Swarm Algorithm}
    \acrodef{UD}{uniform distribution}
    \acrodef{IMA}{iterative multi-agent}
    \acrodef{PDF}{probability density function}

\usepackage[all]{nowidow}
\makeatletter
\g@addto@macro\normalsize{%
  \setlength{\thinmuskip}{2mu}%
  \setlength{\medmuskip}{3mu}%
  \setlength{\thickmuskip}{4mu}%
}
\makeatother

\begin{document}

\let\WriteBookmarks\relax
\def\floatpagepagefraction{1}
\def\textpagefraction{.001}

% \shorttitle{Analysis of MFO}
\shorttitle{}

% \shortauthors{T. Shao, Y. D}
\shortauthors{}

\title [mode = title]{Further Commentary on the Sooty Tern Optimization Algorithm and Tunicate Swarm Algorithm} 



\author{Ngaiming Kwok}
\affiliation{Independent Research}

% \author[1]{Haiyan Shi}[]
% \affiliation[1]{
%     organization={School of Mechanical and Electrical Engineering},
%     addressline={Shaoxing University},
%     state={Zhejiang},
%     city={Shoaoxing},
%     postcode={312000},
%     country={China}}
% \cormark[1]
% \ead{csshy@usx.edu.cn}
% \cortext[1]{Corresponding author}
% \author[2]{Yeping Peng}[]
% \affiliation[2]{
%     organization={Guangdong Key Laboratory of Electromagnetic Control and Intelligent Robots, College of Mechachonics and Control Engineering},
%     addressline={Shenzhen University}, 
%     state={Guangdong},
%     city={Shenzhen}, 
%     postcode={518060},
%     country={China}}
% % \ead{yeping.peng@szu.edu.cn}

\begin{abstract}
In the article \citep{kudela2022commentary}, experimental demonstrations indicated that two \acp{BNIOA}—\ac{STOA} and \ac{TSA}—exhibit a zero-bias, leading to the conclusion that the claims made in the original papers were overstated. In this work, we extend the analysis by investigating the source of this bias from a probabilistic perspective. Our findings suggest that operations involving exponentiation, trigonometric functions, and divisions between random numbers are the primary causes of design flaws. These operations result in probability density distributions with a noticeable shift toward zero. Therefore, the application of these two algorithms should be approached with due caution.
\end{abstract}


% \begin{highlights}
% \item 
% \item 
% \item 
% \end{highlights}

% Keywords
\begin{keywords}
Bio-/Nature-inspired optimization \sep Sooty Tern Optimization Algorithm \sep Tunicate Swarm Algorithm \sep Zero-bias 
\end{keywords}

\maketitle

% \tableofcontents

\section{Introduction}\label{sec:introduction}
\Acfp{BNIOA}, employing the \ac{IMA} approach, have become a vibrant research field in which numerous new algorithms are being proposed regularly. Many of these algorithms are applied to solve optimization problems where derivative-based methods are not applicable.

However, there are also many algorithms that contain fundamental design flaws. One of the primary issues, or even criticisms, is the lack of rigorous theoretical support. This is largely due to the exclusive focus on mimicking the corresponding biological or natural metaphor. Although this approach may replicate the natural process, it does not guarantee that the optimization goal will be achieved. In fact, the choice of metaphor is questionable, as the behavior being mimicked is not necessarily relevant to the intended outcomes. Furthermore, the design of \acp{BNIOA} has become routine, with ideas being reused, combined, and reinterpreted in its metaphor-based framework. 

Several publications in the literature have addressed the aforementioned detrimental problems. For example, insightful critiques are presented in an influential paper \citep{sorensen2015metaheuristics}. Despite these critiques, algorithms with questionable designs continue to be introduced. Another commentary article \citep{kudela2022commentary} also deserves attention. Regrettably, the number of citations it has attracted is orders of magnitude fewer than that of the fallacious algorithms it criticized, the \acf{STOA} \citep{dhiman2019stoa} and \acf{TSA} \citep{kaur2020tunicate} published in \textit{Engineering Applications of Artificial Intelligence}. Therefore, a critical investigation of the root causes of these design flaws is imperative.

The present work is motivated by \citep{kudela2022commentary}. In addition to the experimental demonstration presented there, this study further investigates the design flaws—--specifically, the zero-bias in \ac{STOA} and \ac{TSA}, from a probabilistic perspective. The necessity of maintaining appropriate agent positions according to the \ac{UD} is first asserted from the point of view of random search. The key operations in these two algorithms are reviewed and critically examined. Furthermore, appropriate mathematical reasoning and simulations are provided to illustrate the distortion introduced by the \ac{UD}, which leads to the zero-bias phenomenon. 

\section{Necessity of Uniform Distribution}\label{sec:necessity}
Consider an optimization (without loss of generality, minimization) problem that applies the \ac{IMA} approach, for example:
\begin{equation}\label{eq:minoptimization}
    \min f(\mathbf x),\ \text{such that } f(\mathbf x^*) < f(\mathbf x), 
\end{equation}
where the solution space is $\mathbf x \in {\mathbb R}^D$ of dimension $D$. It is also assumed that the search (feasible) region is bounded within a hyper-volume $V_s \in {\mathbb R}^D$. The \ac{IMA} process can be treated as a sequential process of $T$ iterations that deploy $N$ agents at a time in evaluating the objective function $f(\mathbf x)$. Under these conditions, the probability of finding the optimum at the $t$th-iteration is:
\begin{equation}\label{eq:probfound}
    P(t) = 1 - \left ( 1 - \frac{V_a}{V_s}  \right )^t,
\end{equation}
where $t=1,\cdots,T$ and $V_a \in {\mathbb R}^D$ is the hyper-volume that agent is deployed. It can be seen from Eq. \eqref{eq:probfound} that the probability of finding the optimum is a saturating function. The probability depends on the ratio $V_a/V_s$. If this ratio is increased, then the number of iterations required can be decreased. This is achievable if the search region is fully covered. In order to avoid duplicated search, the agent positions should be distributed uniformly, that is, samples are required to be drawn from the \ac{UD}.

\section{Concerned Algorithms}\label{sec:concerned}
The key computation procedures in the two algorithms, \ac{STOA} and \ac{TSA} are recalled here. The design irrationalities are highlighted. In the following, we use the proper optimization vocabulary instead of in the language of the metaphor. The mathematical expressions are typeset with proper conventions, we distinguish scalars and vectors using italics and bold lowercase, respectively.

\subsection{\acl{STOA}}
This algorithm mimics the migration and attacking behaviors of the sooty terns corresponding to exploration and exploitation respectively.

\paragraph{Exploration: }

\subparagraph{Collision avoidance:}
The position of the agent that does not collide (collision-free) with others is:
\begin{equation}\label{eq:STOA collision}
    \mathbf x_{ca}(t+1) = \mathbf x_{ca}(t) \left ( 2 - \frac{2t}{T} \right ).
\end{equation}

\subparagraph{Convergence to the best-so-far agent:}
In this stage, the agents are driven towards the best-so-far agent $\mathbf x^*(t)$. That is:
\begin{equation}\label{eq:STOA convergence}
    \mathbf x_{cv}(t+1) = 0.5 \mathbf r \otimes (\mathbf x^*(t) - \mathbf x_{cv}(t)),
\end{equation}
where $\mathbf r \sim \mathscr U(0,1)$ with dimensions corresponding to $\mathbf x$ and the symbol $\otimes$ denotes the Hamadard multiplication. 

\subparagraph{Update step-size:}
The agent position update is given by:
\begin{equation}\label{eq:STOA stepsize}
    \Delta \mathbf x(t+1) = \mathbf x_{ca}(t+1) + \mathbf x_{cv}(t+1).
\end{equation}

\paragraph{Exploitation:}

In this phase, agents are driven to follow a spiral trajectory. This is achieved by defining:
\begin{align}\label{eq:STOA trajectory}
    \bm \alpha &= \bm \rho \otimes \sin(\theta),\ \bm \beta = \bm \rho \otimes \cos(\theta),\ \bm \gamma = \bm \rho \otimes \theta,\nonumber \\
    \bm \rho &= \exp(-\theta),\ \theta \sim 2\pi \mathscr U(0,1).
\end{align}
It is important to note that in the original paper, Eq. (9) in \citep{dhiman2019stoa}, $\bm \rho$ is computed from $\exp(\theta)$. The result would be a large number since $\exp(2\pi) = 535.49$. This would drive the agents out of the search space. In the following, the corrected computation is applied.

The updated agent position is obtained from:
\begin{equation}\label{eq:STOA update}
    \mathbf x(t+1) = \mathbf x^*(t) \otimes \Delta \mathbf x(t+1) \otimes (\bm \alpha + \bm \beta + \bm \gamma).
\end{equation}

\subsection{\acl{TSA}}
The \ac{TSA} algorithm mimics the propulsion and swarm behaviors of tunicates. The design incorporates three primary objectives: (1) avoid conflicts, (2) move towards the position of the best-so-far solution, and (3) remain in the vicinity of the best-so-far solution.

\paragraph{Avoiding Conflicts:}
A movement factor used to update the agent positions is given by:
\begin{equation}\label{eq:TSA conflict}
    \mathbf{a} = ({\mathbf r_1 + \mathbf r_2 - 2\mathbf r_3}) \oslash {4\mathbf r_3},
\end{equation}
where $\mathbf r_{1,2,3} \sim \mathscr U(0,1)$ are random variables with dimensions corresponding to $\mathbf{x}$, and the symbol $\oslash$ denotes the Hadamard division.

\paragraph{Movement Towards Best-So-Far:}
The step size for the movement is defined as:
\begin{equation}\label{eq:TSA movement}
    \Delta \mathbf{x}(t) = |\mathbf{x}^*(t) - \mathbf r_4 \otimes \mathbf{x}(t)|,
\end{equation}
where $\mathbf r_4 \sim \mathscr U(0,1)$.

\paragraph{Remaining Around Best-So-Far:}
To keep the agents in the vicinity of the best-so-far agent, the current position is adjusted as follows:
\begin{equation}\label{eq:TSA remain}
    \mathbf{x}(t) \leftarrow
    \begin{cases}
        \mathbf{x}^*(t) + \mathbf{a} \otimes \Delta \mathbf{x}(t), & \text{if } \mathbf r_5 \geq 0.5,\\
        \mathbf{x}^*(t) - \mathbf{a} \otimes \Delta \mathbf{x}(t), & \text{otherwise}.
    \end{cases}
\end{equation}
where $\mathbf r_5 \sim \mathscr U(0,1)$.

\paragraph{Swarm Behavior:}
The final agent position, determined by the swarm behavior, is given by:
\begin{equation}\label{eq:TSA swarm}
    \mathbf{x}(t+1) = \mathbf{x}(t) \oslash (1 + \mathbf r_3).
\end{equation}

\section{Design Flaws and Zero-Bias}

\subsection{\acl{STOA}}
The \ac{STOA} process explicitly makes use of trigonometric functions (with a uniformly distributed parameter $\theta$), see Eq. \eqref{eq:STOA trajectory}. Now consider a random number $\theta \sim \mathscr U(0,2\pi)$. Its \ac{PDF} is $f_\Theta(\theta) = 1/2\pi$. To find the \ac{PDF} of $y=\sin(\theta)$, we transform the variable so that:
\begin{equation}
    f_Y(y) = \sum_i \frac{f_\Theta(\theta_i)}{|dy/d\theta|},
\end{equation}
and the sum is over all possible $\theta$ values such that $y=\sin(\theta)$. There are two solutions for $y \in [-1,+1]$ for $\theta \in [0,2\pi]$. In the interval $[0,\pi]$, $\theta_1 = \arcsin(y)$ and $\theta_2 = \pi - \arcsin(y)$ for the interval $[\pi,2\pi]$. Concerning the derivative, we have $|dy/d\theta| = |\cos(\theta)| = \sqrt{1-y^2}$. The overall \ac{PDF} for $\theta \in [0,2\pi]$ then becomes:
\begin{equation}
    f_Y(y) = \frac{f_\Theta(\theta_1)}{|\cos(\theta_1)|} + \frac{f_\Theta(\theta_2)}{|\cos(\theta_2)|}
    = \frac{1}{\pi \sqrt{1-y^2}}.
\end{equation}
It can be observed that when $y=\pm 1$, the \ac{PDF} tends to infinity and is very distorted from a \ac{UD}. This occurs at $\theta = \pi/2$ and $3\pi/2$ within the interval $[-1,+1]$. A similar procedure would produce the \ac{PDF} for $y=\cos(\theta)$. However, it should be noted that the peaks would occur at $\theta = 0$ and $\pi$.

From Eq. \eqref{eq:STOA update}, the update process contains merely multiplications. For the step-size in \eqref{eq:STOA stepsize}, the first term $\mathbf x_{ca}$ diminishes over iterations. The second term $\mathbf x_{cv}$ would produce a small value due to the produce $0.5 \mathbf r \in [0,0.5]$ in Eq. \eqref{eq:STOA convergence}. Therefore, irrespective of the value of $\mathbf x^*(t)$, the overall update would move towards zero with iterations. This property is not desirable while the location of the optimum cannot be known \textit{a priori} to locate at zero.  

\subsection{\acl{TSA}}
In the conflict-avoidance procedure, as described in Eq. \eqref{eq:TSA conflict}, the update involves divisions of random numbers. For the denominator, there is a non-zero probability that it may be less than one. In such a case, the quotient exceeds the value of the numerator. Conversely, when the denominator exceeds one, the quotient becomes smaller than the numerator. Furthermore, the \ac{PDF} of the quotient will be extended to both the lower and upper bounds of the original denominator, deviating from the \ac{UD}.

In the movement towards the best-so-far solution, as illustrated in Eq. \eqref{eq:TSA movement}, a threshold is set at 0.5. With this setting, the expected value of the modified agent position, $ \mathbf{x}(t) $, will be centered around the best-so-far position, $ \mathbf{x}^*(t) $.

In the imitation of swarm behavior, as outlined in Eq. \eqref{eq:TSA swarm}, division occurs by a shifted random number. Specifically, the denominator, $1 + \mathbf{r}_3$, is greater than one. As a result, the quotient will be smaller than the numerator. In other words, the updated agent position is driven towards zero.

\section{Illustrations of the \ac{PDF}}\label{sec:illustration}
Simulations are conducted to produce the \acp{PDF} of the outcomes from the operations in \ac{STOA} and \ac{TSA}. First, we let the existence of the optimum in the search region be unknown. Furthermore, based on the principle of maximum uncertainty, the \ac{PDF} of the variables follows the \ac{UD} before being operated. It is also assumed that the variable span includes the origin (at zero) of the search space. To obtain smooth \acp{PDF}, $10^6$ samples are used, the span is in $[-100,+100]$, and the resolution is one. Furthermore, the 1-dimensional case is considered where the arithmetic operations, on the elements of a high dimensional agent, are independent.

\begin{figure*}[!h]
    \centering
    \subfigure[]{
    \includegraphics[width=0.275\textwidth]{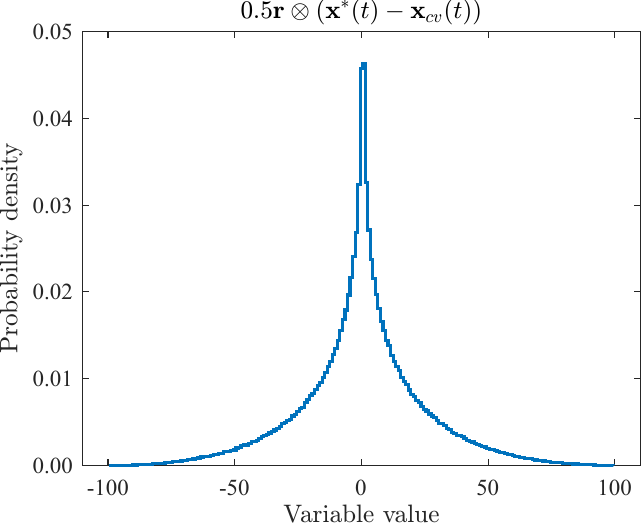}}
    \subfigure[]{
    \includegraphics[width=0.275\textwidth]{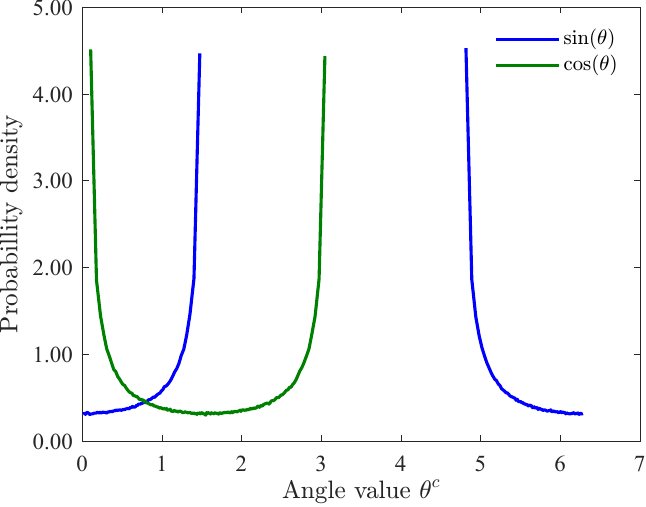}}\\
    \subfigure[]{
    \includegraphics[width=0.275\textwidth]{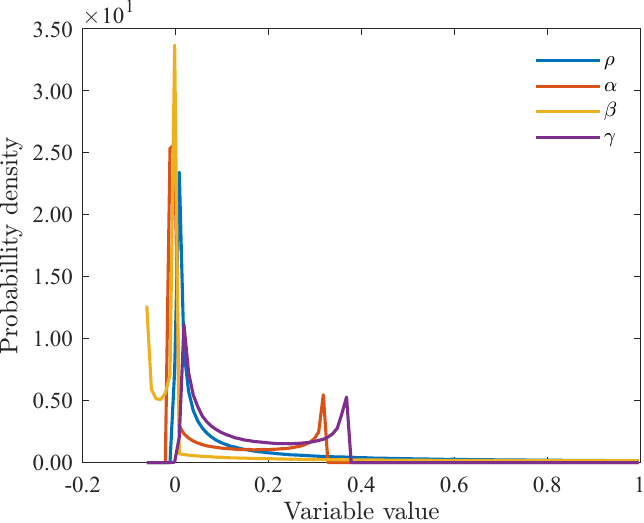}}
    \subfigure[]{
    \includegraphics[width=0.275\textwidth]{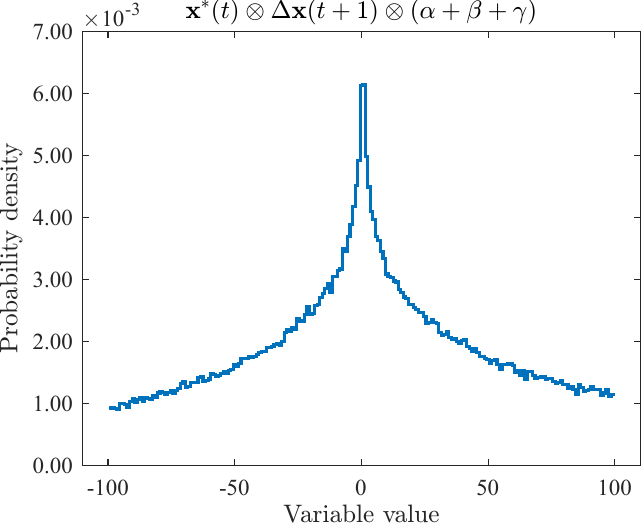}}
    \caption{\Aclp{PDF} of the intermediate \ac{STOA} stages. (a) convergence to best-so-far, (b)sine and cosin terms in Eq. \eqref{eq:STOA trajectory}, (c) exploitation stage, (d) updated agent position.}
    \label{fig:STOA_PDF}
\end{figure*}

\subsection{\acl{STOA}}
The \acp{PDF} of the intermediate processing stages in \ac{STOA} are depicted in Fig. \ref{fig:STOA_PDF}. For the convergence to the best-so-far is illustrated in Eq. \eqref{eq:STOA convergence} and Fig. \ref{fig:STOA_PDF}(a). In the case where $\mathbf{x}^*(t) - \mathbf{x}(t)$ falls within the range $[-1, +1]$, and is multiplied by $\mathbf{r}$ in the interval $[0, 1]$, the product is less than both the multiplier and the multiplicand. As a result, a peak is observed in the \ac{PDF} around zero.

In the exploitation stage, the \acp{PDF} of the $\sin(\theta)$ and $\cos(\theta)$ terms are shown in Fig. \ref{fig:STOA_PDF}(b). Distinct peaks are observed at the angles $0$ or $2\pi$, $\pi/2$, $\pi$, and $3\pi/2$. In this plot, the angle resolution is $10^{-2}$ radians. The \acp{PDF} of $\bm{\alpha}$, $\bm{\beta}$, $\bm{\gamma}$, and $\bm{\rho}$ are presented in Fig. \ref{fig:STOA_PDF}(c). As a correction is applied, the exponent term is bounded such that $\bm{\rho} = \exp(-\theta) \in [0.0019, 1]$. Consequently, the \acp{PDF} are concentrated at small values compared to the range of the search space, which is $[-100, +100]$.

The update procedure described in Eq. \eqref{eq:STOA update} is illustrated by the \ac{PDF} in Fig. \ref{fig:STOA_PDF}(d). Since the update involves the multiplication of three components, the resulting \ac{PDF} exhibits a peak around zero. This suggests that, due to the location of the agents, the algorithm tends to favor optima near zero. This behavior is undesirable, as for problems where the optimum is not near zero, the algorithm may fail to identify the correct optimum.

\begin{figure*}[!h]
    \centering
    \subfigure[]{
    \includegraphics[width=0.275\textwidth]{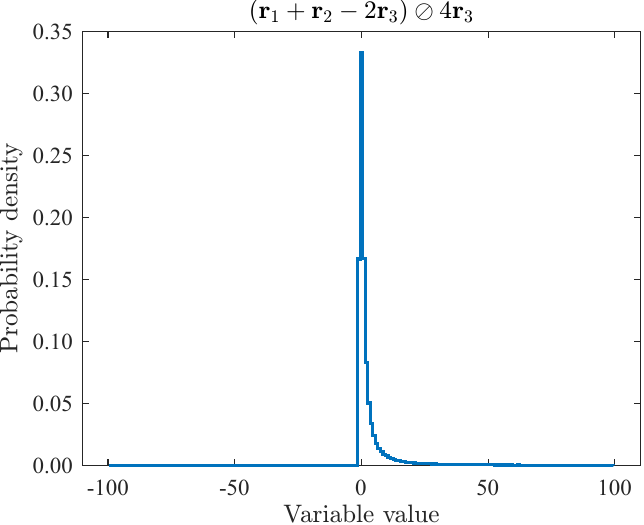}}
    \subfigure[]{
    \includegraphics[width=0.275\textwidth]{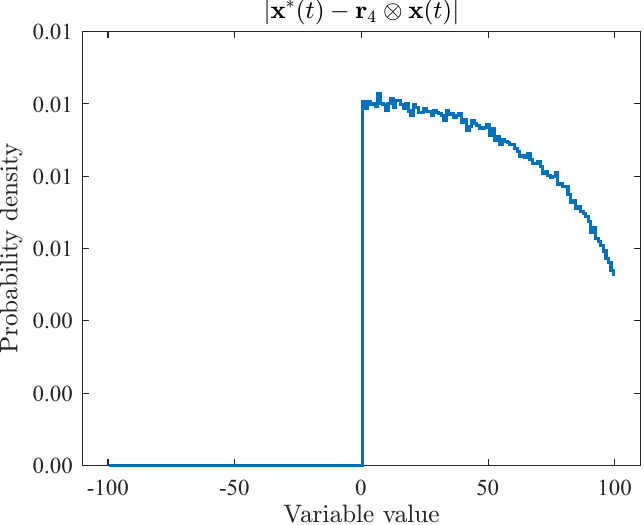}}\\
    \subfigure[]{
    \includegraphics[width=0.275\textwidth]{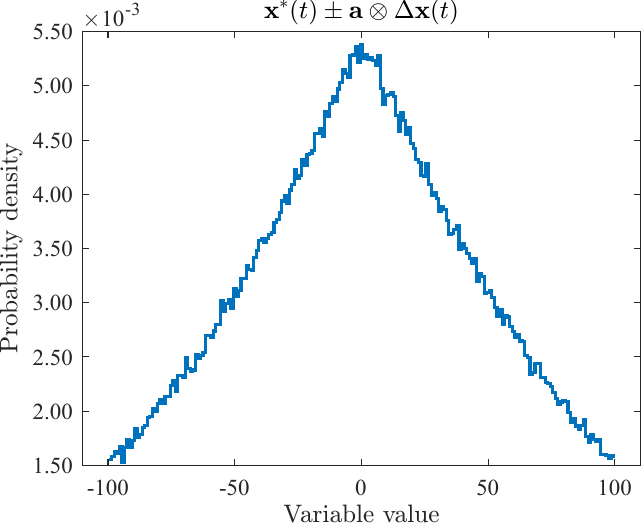}}
    \subfigure[]{
    \includegraphics[width=0.275\textwidth]{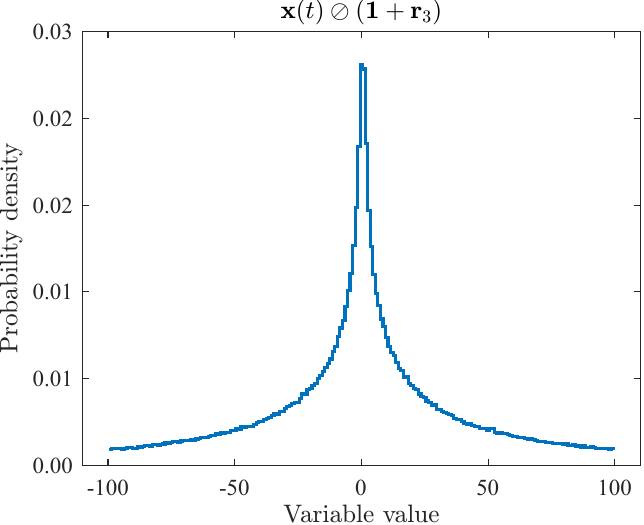}}
    \caption{\Aclp{PDF} of the intermediate \ac{TSA} stages. (a) movement factor $\mathbf a$, (b) movement towards best-so-far, (c) remain around best-so-far, (d) swarm behavior.}
    \label{fig:TSA_PDF}
\end{figure*}

\subsection{\acl{TSA}}
The \acp{PDF} of the results obtained from the different computation stages in \ac{TSA} are illustrated in Fig. \ref{fig:TSA_PDF}. The \ac{PDF} resulting from the computation of the movement factor, as described in Eq. \eqref{eq:TSA movement}, is shown in Fig. \ref{fig:TSA_PDF}(a). A sharp peak is observed near zero. This is primarily due to the fact that the numerator, $4 \mathbf{r}_3$, is larger than the denominator, $\mathbf{r}_1 + \mathbf{r}_2 - 2 \mathbf{r}_3$, causing the quotient to become smaller. Additionally, since both the numerator and the denominator are positive, the quotient is also positive, and the \ac{PDF} does not exist for negative values.

In the movement towards the best-so-far stage, the corresponding \ac{PDF} is shown in Fig. \ref{fig:TSA_PDF}(b). As before, due to the absolute value operation, the \ac{PDF} only exists for positive values. Furthermore, the \ac{PDF} decreases in a power-law manner at higher values.

The \ac{PDF} of the remaining best-so-far stage is depicted in Fig. \ref{fig:TSA_PDF}(c). The \ac{PDF} takes on a triangular shape, with the maximum occurring at zero. The broader peak is the result of a simpler operation, which is merely the difference between a random number and the product of two other random numbers.

Regarding the swarm behavior stage, Fig. \ref{fig:TSA_PDF}(d) illustrates the \ac{PDF}. A narrow spike is observed around zero, which is caused by the division by a random number, $1 + \mathbf{r}_3 > 1$. Consequently, the triangular \ac{PDF} observed in the remaining best-so-far stage is compressed. This phenomenon is undesirable in the search for the optimum.

\subsection{Discussion}
The \acp{PDF} of the final updated agent positions for both the \ac{STOA} and \ac{TSA} exhibit a pronounced peak around zero. This zero-bias phenomenon plays a crucial role in driving a significant portion of the results in each iteration toward zero. As a consequence, unless it is known in advance that the optimum lies near zero, these two algorithms are fundamentally limited in their ability to effectively and efficiently locate the optimum within the search space.

Furthermore, upon examining the computational procedures of these algorithms, it becomes evident that both multiplication and division operations involving random numbers are integral to their functioning. The former is clearly observable in the \ac{STOA} during the update stage, as presented in Eq. \eqref{eq:STOA update}. The latter is likewise apparent in the \ac{TSA}, specifically during the avoiding conflict stage (Eq. \eqref{eq:TSA conflict}) and the swarm behavior stage, see Eq. \eqref{eq:TSA swarm}. In addition to these stages, the overall design of both algorithms incorporates various complex computational steps, the majority of which predominantly involve either multiplication or division. These two arithmetic operations, by their nature, contribute to a reduction in the probability of successfully locating the optimum, as they hinder the uniform distribution of agent positions across the search space. This lack of uniformity significantly impedes the exploration of the entire space, thereby limiting the algorithms' ability to find the true optimum.

In light of these observations, it is essential that future designs of \acp{BNIOA} give careful consideration to the simplification of the operations involved. Moreover, particular attention should be paid to ensuring that the issue of zero-bias can be effectively avoided, thereby enhancing the general applicability of this class of algorithms.

\section{Conclusion}
This paper provides a critical commentary on the \ac{STOA} and \ac{TSA}, directly addressing the theoretical underpinnings of the zero-bias phenomenon highlighted in \citep{kudela2022commentary}. Our analysis, grounded in probability theory, definitively confirms the presence of zero-bias in these algorithms. This is clearly demonstrated by the peaks observed in the \acp{PDF} during various computation stages, including the final update of agent positions. \Acp{BNIOA} that exhibit this significant flaw are fundamentally unsuitable for optimization problems where the optimum is not located at zero. Therefore, any application of the \ac{STOA} and \ac{TSA} should be approached with full awareness of these critical limitations, which severely undermine their effectiveness in broader contexts.

% \section*{Acknowledgement}\label{sec:acknowledgement}
% This work is supported by National Natural Science Foundation of China under Grant No. 52475205, the Natural Science Foundation of Guangdong Province under Grant No. 2024A1515030208, and Shenzhen International Cooperation Research Project under Grant No. GJHZ20220913143005009.  


\balance

\bibliographystyle{cas-model2-names}
\bibliography{CommentarySTOATSA2025BIB}

@article{kudela2022commentary,
  title={Commentary on: “{STOA}: {A} bio-inspired based optimization algorithm for industrial engineering problems ”[{EAAI}, 82 (2019), 148--174] and “{T}unicate Swarm Algorithm: A new bio-inspired based metaheuristic paradigm for global optimization”[{EAAI}, 90 (2020), no. 103541]},
  author={Kudela, Jakub},
  journal={Engineering Applications of Artificial Intelligence},
  volume={113},
  pages={104930},
  year={2022},
  publisher={Elsevier}
}

@article{dhiman2019stoa,
  title={{STOA}: {A} bio-inspired based optimization algorithm for industrial engineering problems},
  author={Dhiman, Gaurav and Kaur, Amandeep},
  journal={Engineering Applications of Artificial Intelligence},
  volume={82},
  pages={148--174},
  year={2019},
  publisher={Elsevier}
}

@article{kaur2020tunicate,
  title={Tunicate Swarm Algorithm: {A} new bio-inspired based metaheuristic paradigm for global optimization},
  author={Kaur, Satnam and Awasthi, Lalit K and Sangal, Amrit Lal and Dhiman, Gaurav},
  journal={Engineering Applications of Artificial Intelligence},
  volume={90},
  pages={103541},
  year={2020},
  publisher={Elsevier}
}

@article{sorensen2015metaheuristics,
  title={Metaheuristics—the metaphor exposed},
  author={S{\"o}rensen, Kenneth},
  journal={International Transactions in Operational Research},
  volume={22},
  number={1},
  pages={3--18},
  year={2015},
  publisher={Wiley Online Library}
}

% \balance

\end{document}